\documentclass[conference]{IEEEtran}
\IEEEoverridecommandlockouts
\usepackage{cite}
\usepackage{amsmath,amssymb,amsfonts}
\usepackage{algorithmic}
\usepackage{graphicx}
\usepackage{textcomp}
\usepackage{xcolor}
\usepackage{booktabs}
\usepackage{svg}
\usepackage{graphicx}
\usepackage{enumitem}
\usepackage{subcaption}
\usepackage{graphicx}

\def\BibTeX{{\rm B\kern-.05em{\sc i\kern-.025em b}\kern-.08em
    T\kern-.1667em\lower.7ex\hbox{E}\kern-.125emX}}

\IEEEaftertitletext{\vspace{-1\baselineskip}}
\begin{document}

\title{Towards Improved Illicit Node Detection  \\ 
with Positive-Unlabelled Learning\\
}


\author{\IEEEauthorblockN{Junliang Luo\IEEEauthorrefmark{1},
Farimah Poursafaei\IEEEauthorrefmark{2} and Xue Liu\IEEEauthorrefmark{3}}
\IEEEauthorblockA{School of Computer Science, McGill University, Montréal, Québec, Canada\\
 \IEEEauthorrefmark{1}junliang.luo@mail.mcgill.ca,
\IEEEauthorrefmark{2}farimah.ramezanpoursafaei@mail.mcgill.ca,
\IEEEauthorrefmark{3}xueliu@cs.mcgill.ca}}

\IEEEoverridecommandlockouts

\IEEEpubid{\makebox[\columnwidth]{979-8-3503-1019-1/23/\$31.00~\copyright2023 IEEE \hfill} \hspace{\columnsep}\makebox[\columnwidth]{ }}

\maketitle

\begin{abstract}
Detecting illicit nodes on blockchain networks is a valuable task for strengthening future regulation.
Recent machine learning-based methods proposed to tackle the tasks are using some blockchain transaction datasets with a small portion of samples labeled positive and the rest unlabelled (PU).
Albeit the assumption that a random sample of unlabeled nodes are normal nodes is used in some works, we discuss that the label mechanism assumption for the hidden positive labels, and its effect on the evaluation metrics is worth considering.
We further explore that PU classifiers dealing with potential hidden positive labels can have improved performance compared to regular machine learning models.
We test the PU classifiers with a list of graph representation learning methods for obtaining different feature distributions for the same data to have more reliable results.
\end{abstract}


\section{Introduction}
The nature of anonymity and decentralization of blockchain systems are making changes in the finance industry due to its immutability, transparency, and automation \cite{schar2021decentralized}.
Such decentralized systems, however, are in the current stage of a temporarily unregulated environment \cite{barbereau2022decentralised, makarov2022cryptocurrencies} with a variety of abnormal usages and security concerns.
%
The abnormal usages include both the illicit activities clearly defined by traditional fiance: phishing scams, Ponzi schemes, money laundering, etc. \cite{cumming2015financial}, and also the activities with no clear definition of lawfulness or being just defined such as mixing services, i.e., the mixer nodes involve in funds to confuse the trace of the transfers from the original source, e.g., US Department of the Treasury declared Tornado Cash as a sanctioned entity \cite{anthony2022treasury, US_treasury_2022}.
To strengthen the future regulation of the decentralized financial markets, the detection of addresses that performed illicit activities helps in assessing security issues.
%
%
Research is conducted using machine learning models on blockchain transaction data for detecting illicit nodes (addresses).
The methods include using topological graph features as the input to machine learning models such as tree-based models \cite{sun2019regulating}, clustering\cite{ofori2021topological, akcora2020bitcoinheist}, logistic regression \cite{wu2021detecting},
and using the node embedding vectors extracted by graph representation learning as the input to perform node classification \cite{liu2019hyperbolic, alarab2020competence, martin2022anomaly, poursafaei2021sigtran}.
%


%
Whilst much work has been done proposing graph learning models for illicit node classification and reporting the performance by standard classification metrics, little has been discussed about label assumptions or the evaluation methods for the blockchain datasets with illicit node labels such as \cite{wu2020phishers, weber2019anti}, which consist of only very few positive labeled samples and a large portion of unlabelled samples.
Some previous works on detecting phishing nodes use binary labels: phishing and normal nodes with the assumption that the unlabeled nodes are normal  \cite{wu2020phishers, poursafaei2021sigtran, chen2020phishing}.
The assumption is practicable if the majority of nodes are not illicit.
Nevertheless, standard classification metrics such as precision and F1 score cannot be precisely calculated without fully labeled data.
Therefore, the labeling mechanism assumption for the limited positive labels, the effect of the assumption on the evaluation metrics, and the learning methods on those PU data are worth being discussed for providing convincing estimations for the performance.
In this work, we combine existing studies of PU learning on illicit node classification. 
We first discuss the label mechanism assumption of the Completely At Random (SCAR) assumption is necessary for reducing the PU learning to standard binary classification.
We demonstrate the difference between the estimated values of evaluation metrics and the actual values through an engineered PU dataset from the Ethereum transaction dataset proposed in \cite{wu2020phishers} to show the concerns of assuming unlabeled data to be normal.
We conduct experiments to show that applying various PU classifiers can help in improving the classification performance on two real-world datasets with limited positive labels.
The PU classifiers estimate potential identifiable class prior or treat the unlabeled examples as negative samples with label noise and learn with biased models.
We also compare various graph representation methods for extracting node embedding vectors as the input to get diverse data distribution for the same data to obtain more reliable results.
The results show that PU classifiers make improvements on the results compared to the baselines.
The result evidence that PU learning help in learning the data with possible hidden positive nodes, which will be a common situation in the blockchain transaction data.

\section{Related work}
Related research related to our work mostly includes node classification tasks in blockchain transaction networks using machine learning models.
The tasks include phishing and scam nodes detection \cite{poursafaei2021sigtran}, mixing service nodes detection \cite{wu2021detecting}, and blockchain users deanonymizing \cite{beres2021blockchain}.
%
%
%
Among the node classification approaches on blockchain transaction data, a few works apply PU learning models \cite{wu2021detecting,jin2022detecting}.
In particular, Wu et al. \cite{wu2021detecting} use the two-step spy method to select some reliable negative instances from the unlabeled data for better detecting the nodes of mixers in Bitcoin networks.
%
%
However, little discussion has been devoted to the label assumption and investigation of the data.
Saunders et al. \cite{saunders2022evaluating} recently discuss the proper evaluation approaches and argue that the label mechanism assumption is essential for obtaining convincing results for the PU datasets.

\section{Positive and unlabeled (PU) learning}
\label{positive_and_unlabeled(PU)_learning}
The major challenge for applying machine learning models on blockchain transaction networks is the lack of labelled data \cite{venkatesh2020blockchaining}.
As has also been succinctly stated in the aforementioned sections, unlabeled addresses can be negative or actually hidden positive, since the label resource being commonly used are from block explorer and analytic platforms such as Etherscan Label Word Cloud \cite{label_word_cloud_etherscan} for Ethereum. 
The labels will keep being collected from the public submissions, so some unlabeled addresses will become labeled in the future.
The node classification tasks on blockchain transaction data are very likely to be appropriate to the PU circumstances.

\subsection{Selected Completely At Random (SCAR) Assumption }
When modeling blockchain transaction illicit node detection as a binary classification task, a PU dataset has limited number of positive nodes labeled as +1 and unlabeled nodes labeled -1 or 0, meaning that none of the negative examples are labeled.
To reduce a PU task to a standard binary classification, the data must be under Selected Completely At Random (SCAR) assumption \cite{elkan2008learning}: the labeled set is selected independently from its attributes, i.e., the positive samples are identically distributed samples from the positive data distribution. 
Otherwise, all the standard classification metrics cannot be estimated.
SCAR assumption is defined as below using the same notation in \cite{bekker2020learning}, where $y$ is the true label most data missing, $s$ is the existing label for the labeled sample data, and $c$ is a constant for label frequency.
\begin{equation}
    Pr(s=1 | x, y = 1) = Pr(s=1 | y=1) = c
\end{equation}   
Let $f(x)$ be a classifier that predicts $Pr(y=1|x)$, and $g(x)$ be a classifier that predicts $Pr(s=1|x)$:
\begin{equation}
    c \cdot f(x) = g(x)
    \label{eqn:SCAR_classifier}
\end{equation}  
Elkan et al. \cite{elkan2008learning} mention that predicting
$Pr(y=1|x)$ is different from predicting $Pr(s=1|x)$ by a constant under SCAR assumption, which means if both classifiers rank instances by their predicted probability, the data samples with the highest predicted positive probability will be the same for both $g(x)$ and $f(x)$, which means recall can be estimated.

\subsection{Evaluation metrics}
\vspace{-0.17cm}
\begin{figure}[!htbp]
    \centering
    \includegraphics[width=0.31\textwidth, scale=1]{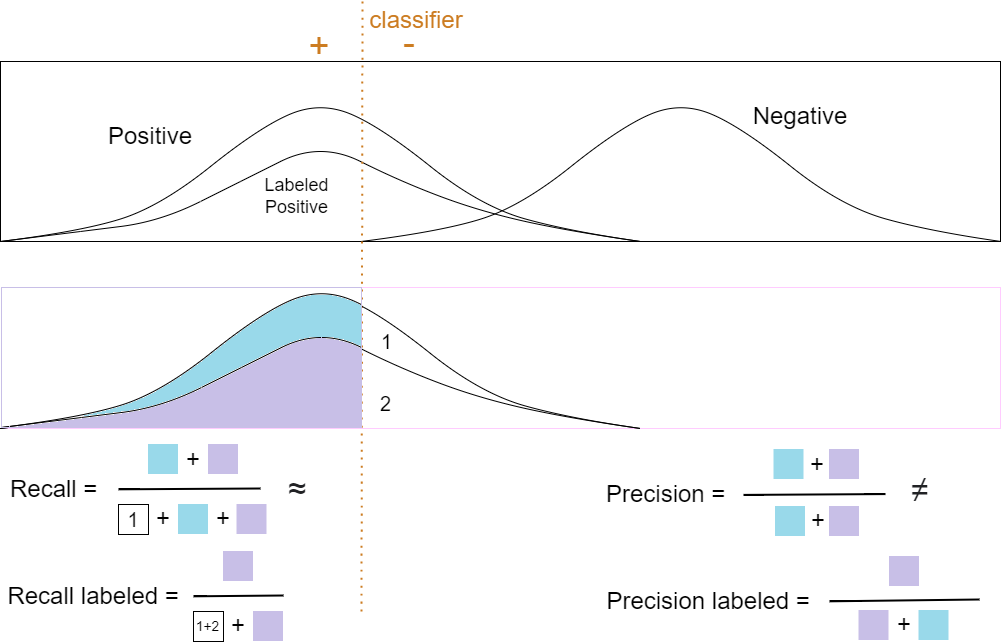} 
    \caption{Under SCAR assumption, the labeled positives are selected uniformly at random from the true positives.}
    \label{fig:label_mechanism_assumption_SCAR}
\end{figure}

\begin{figure}[!htbp]
    \centering
    \includegraphics[width=0.3796\textwidth, scale=1]{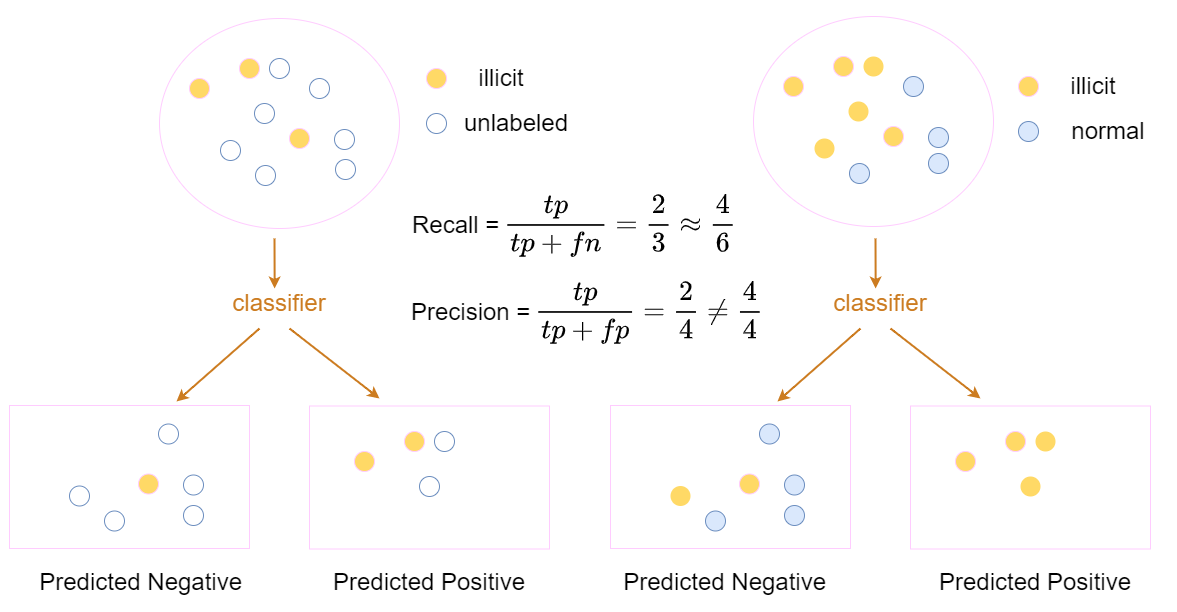} 
    \caption{Example of node classification on PU data under SCAR assumption. Right shows unlabeled nodes with true labels.}
    \label{fig:PU_example_clr}
\end{figure}

We present an example of a PU dataset under the SCAR assumption and classification results on an example PU data in Figure \ref{fig:label_mechanism_assumption_SCAR} and \ref{fig:PU_example_clr}.
Under the SCAR assumption, the recall $\text{r}=Pr(\hat{y}=1|y=1)$ can be estimated from PU data $\text{r}=Pr(\hat{y}=1|s=1)$ due to the property mentioned in Equation \ref{eqn:SCAR_classifier}, but precision cannot be calculated, since the false positive cannot be correctly estimated as illustrated in Figure. \ref{fig:PU_example_clr}.

However, some previous works such as \cite{wu2020phishers, poursafaei2021sigtran} mention that the data is unlabeled and positive but do not mention any assumptions regarding the negative label mechanism assumption. 
The presented evaluation results of precision, recall, and F1-score is based on assuming the unlabeled is negative.
The results are reliable under their experiment setting of randomly selecting the same number 
of unlabeled nodes along with the positive nodes to be used as a dataset.
The chance of picking up an unlabeled positive (unlabeled illicit) node is much smaller than picking up a negative (normal) node.
However, without noting that the results are an
estimation based on the SCAR assumption could raise the concern about similar future works on blockchain transaction data which is likely to be positive and unlabeled.
Since the prediction $\text{p} = Pr(y=1 | \hat{y}=1)$ cannot be estimated from PU data, the F1 score which is defined as the harmonic mean of precision and recall, $\text{F1} = 2pr / (p+r) $, cannot be defined either. 
As an alternative of F1 score, $\text{PUF1} = r^2 / Pr(\hat{y}=1)$ is proposed and suggested in \cite{lee2003learning, bekker2020learning}, replacing the precision by Bayesian inference in a F1-similar measure combing precision and recall $pr / Pr(y=1) $ by multiplying both numerator and denominator with a recall: $pr^2 / r \cdot Pr(y=1) = Pr(y=1|\hat{y}=1)r^2 / Pr(\hat{y}=1 |, y=1) = r^2 / Pr(\hat{y}=1)$. 
The PUF1 will be of a large value when a recall is large but gets punished if the percentage of the positive prediction is large.

\subsection{Evaluation Metrics on Engineered PU Data}
\label{experiment_engineered_PU_data}
%
%
To simulate the estimation compared to the true values of the evaluation metrics on PU data learning assuming the existing unlabeled as negative, we engineer a PU dataset by making some known positive samples as negative (unlabeled).
The dataset we use is a recent frequently used Ethereum transactions dataset presented in \cite{wu2020phishers}, which has \textit{1165} labeled illicit nodes in \textit{2.9} million nodes.
Our experimental setup follows the setting presented in \cite{wu2020phishers, poursafaei2021sigtran}.
We randomly pick the same number of \textit{1165} unlabeled nodes as negative nodes and extracting the first-order neighbors of the positive nodes and the negative nodes and all the connected edges between them to form a subnetwork.
We repeat the random selection ten times to get ten subnetworks where the average number of nodes of the subnetworks is \textit{30,000} nodes and \textit{263,000} edges.
We report the average results for the evaluation.
After randomly selecting some positive labeled nodes to be labeled as negatives, the data was split into 80\% for training and 20\% for testing.
We use node2vec \cite{grover2016node2vec} (detailed explained in Section \ref{node_embedding_learning}) to extract the node embeddings for each subnetwork.
We exploit a Logistic Regression and an SVM model as two frequently used machine learning models for blockchain node classification \cite{khan2022graph}.

\begin{figure}[!htbp]
    \centering
    \includegraphics[width=0.37\textwidth, scale=1]{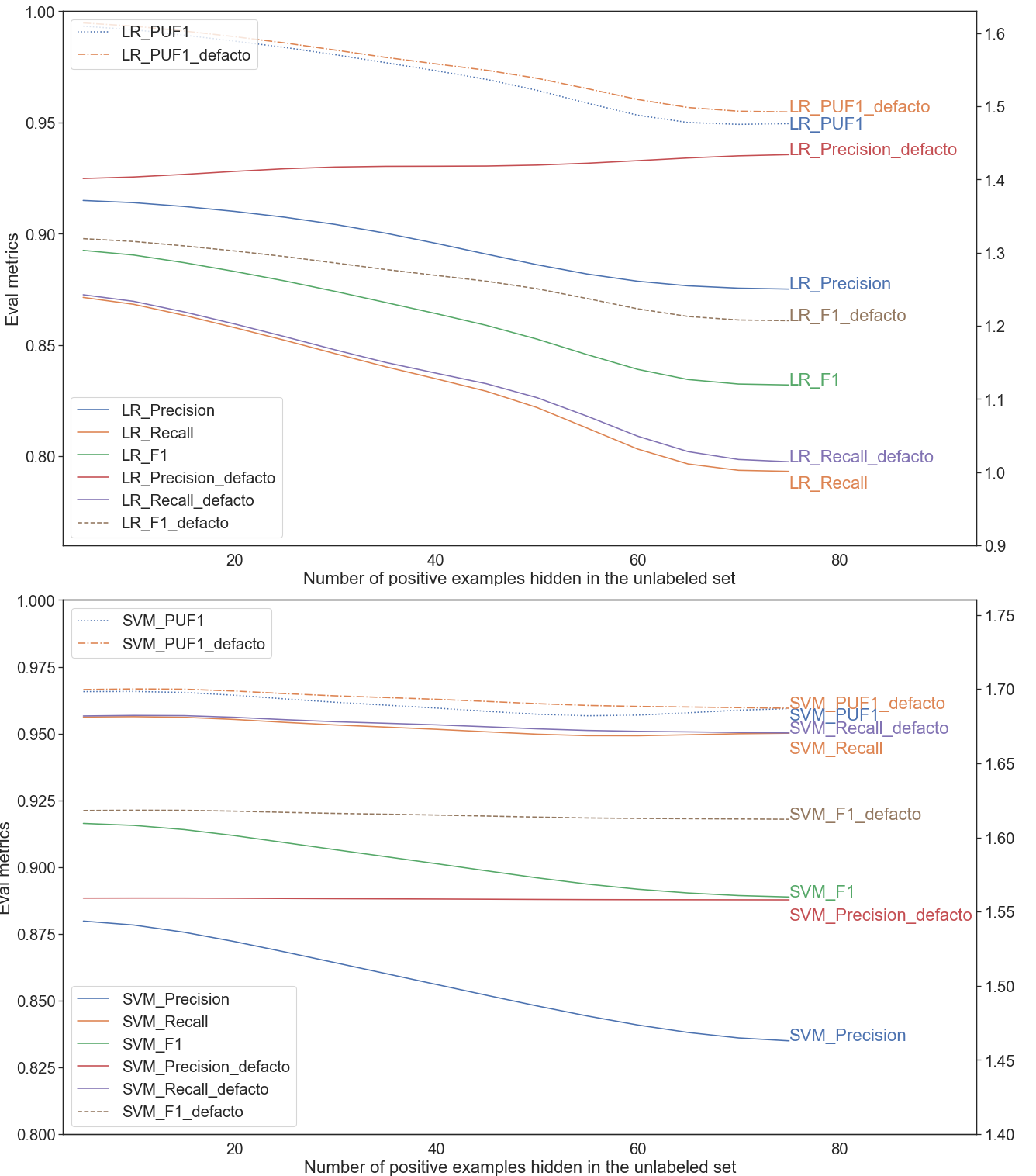} 
    \caption{Results of evaluation metrics against the increasing number of positive nodes hidden in the unlabeled set for the Logistic regression model (top) and SVM model (bottom).}
    \label{fig:engineering_PU_data_experiment}
\end{figure}
All the hidden positives in the training sets are labeled as negative, while in the testing set, the results of the estimated evaluation are calculated by treating the hidden positives as negatives, and the defacto evaluation results treat the hidden positives as positives.
Figure. \ref{fig:engineering_PU_data_experiment} demonstrated different results against the increasing number of hidden positives in the dataset.
The gap between the precision and precision defacto, the gap between F1 and F1 defacto, and the much smaller gap between recall and recall defacto confirm the rationality of the SCAR assumption on the Ethereum transaction dataset when the number of hidden positives is not large.
The percentage of the detected positives in a subset of positives is an estimate for the whole positive set, while some positives labeled as negative (unlabeled) are correctly predicted but counted as false positives, therefore the precision decreases.
In illicit detection or any other similar minority anomalies node
detection, the labeled positives are smaller than the reality and the other class being unlabeled makes the true positives underestimated and the false positives overestimated.
Besides, when the cost of a false negative is higher than the cost of a false positive, i.e., the tolerance is small for miss-detecting known positives, we can treat recall as a more important metric to maximize. However, we must ensure that the positive predictions $Pr(\hat{y}=1)$ are not too large.
In this case, the PUF1 metric defined in Section III-B can be a suitable alternative metric for F1 for the PU data.

\begin{table*}[!htbp]
\renewcommand{\arraystretch}{1.26}
\huge
\caption{Evaluation results on list of ML models and graph embeddings on Ethereum dataset }\label{tab:total_table}
\centering
\resizebox{0.99\textwidth}{!}{%
\begin{tabular}{@{}ccccccccccccccccccccccccccccc@{}}
\toprule
 &  & \multicolumn{6}{c}{Precision} &  & \multicolumn{6}{c}{Recall} &  & \multicolumn{6}{c}{F1} &  & \multicolumn{6}{c}{PUF1} \\ \midrule
 &  & LR & RF & SVM & BA & ET & UPU &  & LR & RF & SVM & BA & ET & UPU &  & LR & RF & SVM & BA & ET & UPU &  & LR & RF & SVM & BA & ET & UPU \\ \cmidrule(r){1-8} \cmidrule(lr){10-15} \cmidrule(lr){17-22} \cmidrule(l){24-29} 
Node2vec &  & 0.937 & 0.913 & 0.904 & 0.904 & 0.909 & 0.849 &  & 0.893 & 0.967 & 0.971 & 0.97 & 0.968 & 0.983 &  & 0.915 & 0.938 & 0.936 & 0.936 & \textbf{0.938} & 0.911 &  & 1.674 & \textbf{1.766} & 1.756 & 1.755 & 1.761 & 1.67 \\
Poincaré &  & 0.628 & 0.676 & 0.738 & 0.717 & 0.685 & 0.574 &  & 0.66 & 0.88 & 0.788 & 0.852 & 0.903 & 0.886 &  & 0.644 & 0.765 & 0.762 & 0.779 & \textbf{0.779} & 0.695 &  & 0.83 & 1.19 & 1.163 & 1.222 & \textbf{1.236} & 1.015 \\
Role2vec &  & 0.912 & 0.899 & 0.928 & 0.915 & 0.913 & 0.659 &  & 0.879 & 0.887 & 0.953 & 0.961 & 0.958 & 0.979 &  & 0.895 & 0.893 & \textbf{0.94} & 0.937 & 0.934 & 0.787 &  & 1.603 & 1.595 & \textbf{1.769} & 1.759 & 1.748 & 1.289 \\
MNMF &  & 0.572 & 0.85 & 0.66 & 0.655 & 0.501 & 0.552 &  & 0.487 & 0.569 & 0.709 & 0.713 & 0.99 & 0.852 &  & 0.523 & 0.681 & 0.683 & \textbf{0.683} & 0.666 & 0.67 &  & 0.555 & 0.966 & 0.936 & 0.934 & \textbf{0.993} & 0.94 \\
Node2vec $\bigoplus$ Poincaré &  & 0.942 & 0.912 & 0.939 & 0.932 & 0.929 & 0.802 &  & 0.909 & 0.969 & 0.946 & 0.96 & 0.952 & 0.983 &  & 0.925 & 0.94 & 0.943 & \textbf{0.946} & 0.94 & 0.883 &  & 1.713 & 1.768 & 1.777 & \textbf{1.789} & 1.768 & 1.577 \\
Node2vec $\bigoplus$ MNMF &  & 0.939 & 0.91 & 0.922 & 0.907 & 0.908 & 0.837 &  & 0.899 & 0.967 & 0.947 & 0.974 & 0.972 & 0.983 &  & 0.918 & 0.938 & 0.934 & \textbf{0.94} & 0.939 & 1.644 &  & 1.688 & 1.762 & 1.746 & \textbf{1.768} & 1.766 & 0.904 \\ \bottomrule
\end{tabular}
}
\label{eth_PU_evaluation_results}
\end{table*}

\begin{table*}[!htbp]
\renewcommand{\arraystretch}{1.26}
\huge
\caption{Evaluation results on list of ML models and graph embeddings on Bitcoin dataset }\label{tab:total_table}
\centering
\resizebox{0.99\textwidth}{!}{%
\begin{tabular}{@{}ccccccccccccccccccccccccccccc@{}}
\toprule
 &  & \multicolumn{6}{c}{Precision} &  & \multicolumn{6}{c}{Recall} &  & \multicolumn{6}{c}{F1} &  & \multicolumn{6}{c}{PUF1} \\ \midrule
 &  & LR & RF & SVM & BA & ET & UPU &  & LR & RF & SVM & BA & ET & UPU &  & LR & RF & SVM & BA & ET & UPU &  & LR & RF & SVM & BA & ET & UPU \\ \cmidrule(r){1-8} \cmidrule(lr){10-15} \cmidrule(lr){17-22} \cmidrule(l){24-29} 
Node2vec &  & 0.632 & 0.639 & 0.656 & 0.654 & 0.652 & 0.546 &  & 0.77 & 0.816 & 0.928 & 0.93 & 0.936 & 0.969 &  & 0.694 & 0.717 & \textbf{0.769} & 0.768 & 0.768 & 0.699 &  & 0.974 & 1.043 & 1.218 & 1.216 & \textbf{1.22} & 1.059 \\
Poincaré &  & 0.544 & 0.612 & 0.678 & 0.638 & 0.536 & 0.597 &  & 0.557 & 0.541 & 0.700 & 0.769 & 0.938 & 0.646 &  & 0.55 & 0.574 & 0.689 & \textbf{0.698} & 0.683 & 0.539 &  & 0.606 & 0.662 & 0.949 & 0.982 & \textbf{1.007} & 0.703 \\
Role2vec &  & 0.655 & 0.703 & 0.745 & 0.729 & 0.671 & 0.541 &  & 0.695 & 0.724 & 0.800 & 0.824 & 0.885 & 0.915 &  & 0.675 & 0.713 & 0.772 & \textbf{0.774} & 0.763 & 0.68 &  & 0.911 & 1.018 & 1.192 & \textbf{1.202} & 1.187 & 0.99 \\
MNMF &  & 0.601 & 0.589 & 0.645 & 0.645 & 0.548 & 0.540 &  & 0.477 & 0.838 & 0.776 & 0.786 & 0.978 & 0.730 &  & 0.531 & 0.691 & 0.704 & \textbf{0.708} & 0.703 & 0.590 &  & 0.574 & 0.987 & 1.000 & 1.013 & \textbf{1.073} & 0.768 \\
Node2vec $\bigoplus$ Poincaré &  & 0.624 & 0.636 & 0.757 & 0.72 & 0.697 & 0.552 &  & 0.742 & 0.795 & 0.813 & 0.87 & 0.891 & 0.973 &  & 0.678 & 0.707 & 0.784 & \textbf{0.788} & 0.782 & 0.704 &  & 0.927 & 1.012 & 1.231 & \textbf{1.253} & 1.242 & 1.074 \\
Node2vec $\bigoplus$ MNMF &  & 0.625 & 0.635 & 0.716 & 0.708 & 0.689 & 0.547 &  & 0.717 & 0.796 & 0.873 & 0.886 & 0.908 & 0.975 &  & 0.668 & 0.707 & 0.787 & \textbf{0.787} & 0.783 & 0.701 &  & 0.897 & 1.011 & 1.25 & \textbf{1.256} & 1.251 & 1.067 \\ \bottomrule
\end{tabular}
}
\label{btc_PU_evaluation_results}
\end{table*}

\section{Node embeddings and PU learning models}
The previous experiment was engineering a PU dataset
to demonstrate that the evaluations are an estimation of some defacto evaluations which are unknowable on the current set of positive labels.
However, even using all the current collected positive labels, the dataset itself is a genuine PU dataset where there must be hidden positive labels missing, and very likely more positive labels can be added in the future for some existing nodes in the dataset. 
In this section, we investigate whether some widely used PU models can be utilized for a better illicit node classification performance on two real-world transaction datasets, even though the precision and F1 may be underestimated due to hidden positive labels, the comparison between PU models is still fair.
The PU models that we focus on mostly consider the unlabeled class as a negative class but with class label noise given the SCAR assumption.
The datasets are the Ethereum transaction dataset \cite{wu2020phishers} with \textit{1165} nodes labeled illicit in \textit{2.9} million nodes and also a Bitcoin transaction dataset (Elliptic dataset) with \textit{4545} illicit nodes labeled in \textit{203,000} nodes of Bitcoin networks \cite{weber2019anti}.

\subsection{Node Embedding}
\label{node_embedding_learning}
The input to PU learning models is the node embedding vectors, which can be extracted by various graph representation learning algorithms.
A graph learning algorithm maps the nodes or edges of a graph into the low dimensional vectors that represent certain features depending on which the algorithm emphasizes.
These vectors cost less space to store and less computation power to perform the downstream learning task.
The feature vectors are given as inputs to the PU machine learning models for the supervised classification with the node labels.
We list the following graph representation algorithms for extracting the node embedding vectors, which emphasize the features of neighbors, hierarchies, structures, and communities, with an output dimension of 64 for all the methods.

\begin{itemize}
    \item \textit{Node2vec} \cite{grover2016node2vec} is a random-walk-based node representation method that can preserve the node neighbor relations in the embedding space. The similarity function is defined as the probability for the node $u$ to reach node $v$ on a random walk of a certain length. We use \textit{$\text{p=1}, \text{q=1}$} to not add bias moving  forward or return from the current node. We use walk length \textit{$\text{l} = 5$} inline with \cite{poursafaei2021sigtran}. 
    \item \textit{Poincaré} \cite{NIPS2017_59dfa2df} is a hierarchical representation learning method that is based on Riemannian optimization and learns node embedding in a hyperbolic Poincaré space which can better capture the hierarchy information of nodes. The nodes in the high level of the hierarchy are the nodes that connect to many other nodes which are mostly not connected themselves, such as the whale nodes \cite{liu2019hyperbolic}. The higher the hierarchy, the closer the node is to the coordinate origin.
    \item \textit{Role2vec} \cite{ahmed2018learning} is a random-walk-based structural node embedding method (with walk length $\text{l = 5}$).
    Here, the two addresses that have similar motif structure in the transaction graph are considered to be more similar despite their distance.
    \item \textit{MNMF} \cite{wang2017community} is a matrix factorization-based method that learns a cluster membership distribution over nodes such that a node’s embedding preserves both itself and community structures.
\end{itemize}

\begin{figure}[!htbp]
\centering
  \begin{subfigure}[t]{.197\textwidth}
    \centering
    \includegraphics[width=\linewidth]{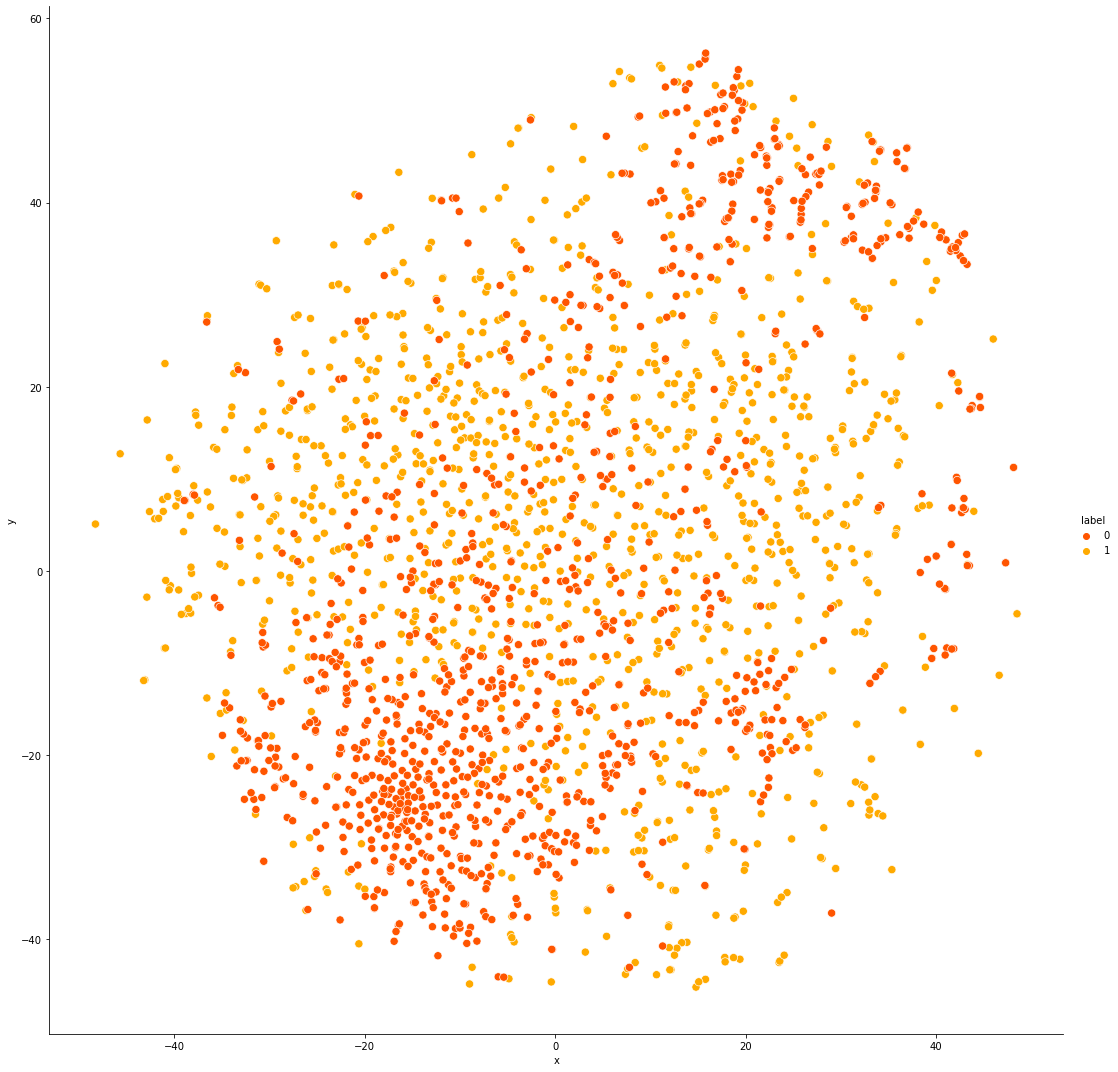}
    \caption{Node2vec}
  \end{subfigure}
  \begin{subfigure}[t]{.197\textwidth}
    \centering
    \includegraphics[width=\linewidth]{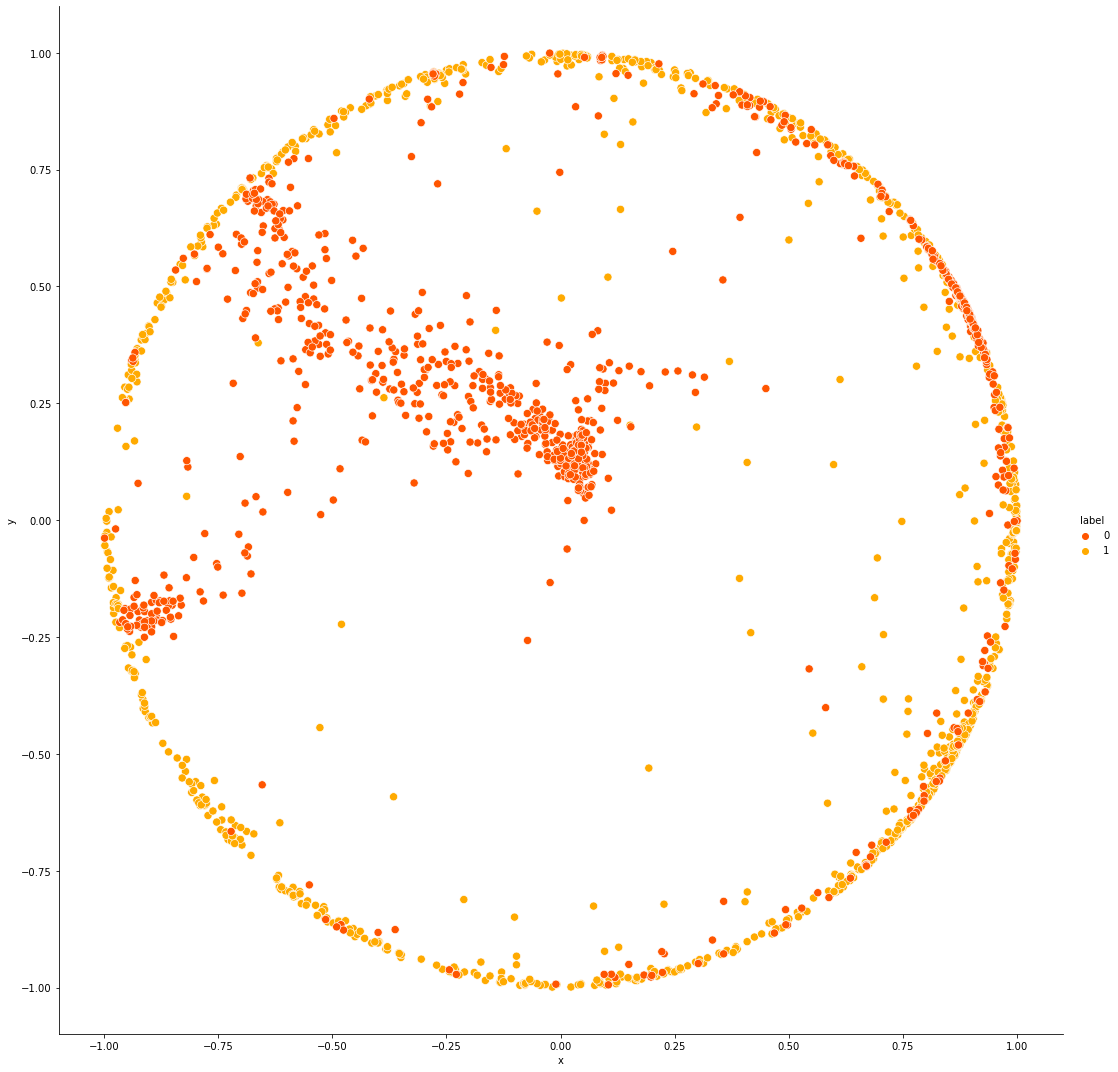}
    \caption{Poincaré}
  \end{subfigure}

  \medskip

    \centering
  \begin{subfigure}[t]{.197\textwidth}
    \centering
    \includegraphics[width=\linewidth]{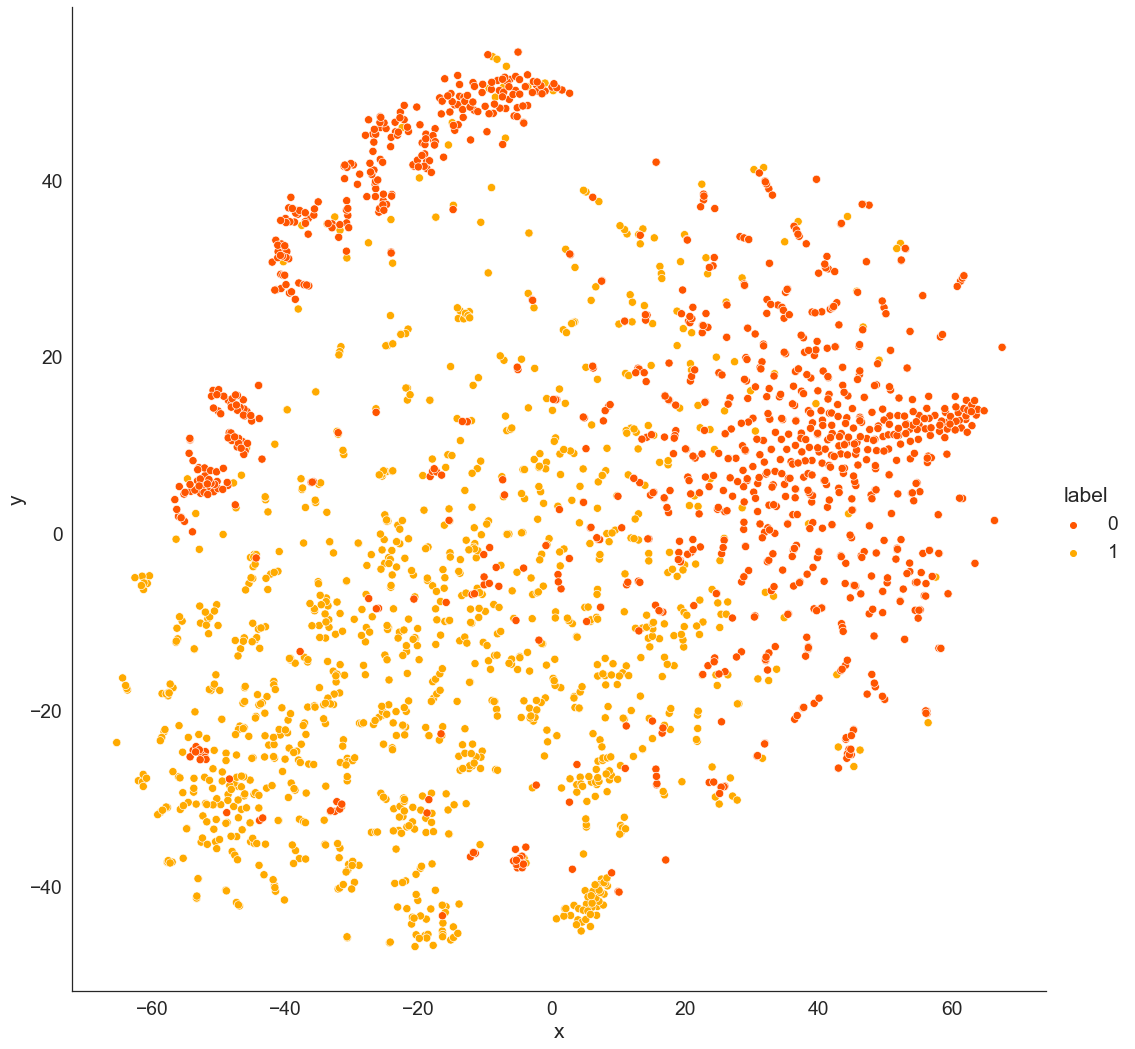}
    \caption{Role2Vec}
  \end{subfigure}
  \begin{subfigure}[t]{.197\textwidth}
    \centering
    \includegraphics[width=\linewidth]{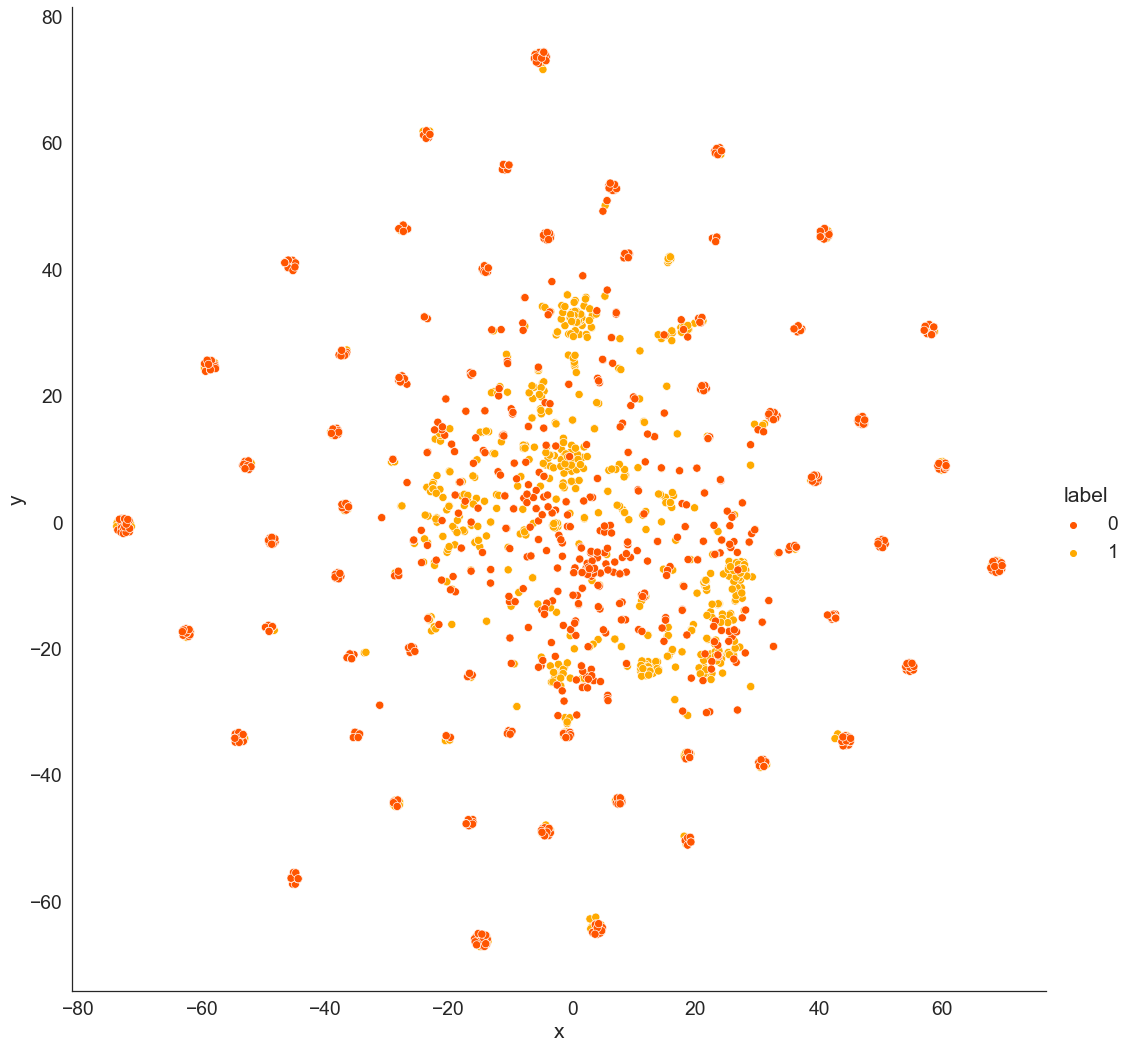}
    \caption{MNMF}
  \end{subfigure}
  \caption{Visualization (t-SNE dimension reduced to 2D) of the extracted node embedding vectors using graph representation models on a subgraph of Ethereum transaction data.}
\end{figure}


%
We use Logistic Regression (LR), Random Forest (RF), and Support Vector Machine (SVM) as the baselines.
In addition, we explore PU learning models, including the bagging (BAG)-SVM \cite{mordelet2014bagging} (which learns and aggregates multiple biased SVM classifiers trained on the positive samples and a subset of the negative samples to discriminate positive samples with the small subset of negative samples), Elkanoto (EL)-SVM \cite{elkan2008learning} (which trains a classifier to predict $P(s=1 | x)$, and use the classifier to estimate the probability of positive samples $P(s=1|y=1)$ are labeled, then the probability of a data $k$ is positive is $P(s=1|k) / P(s=1|y=1)$), and Unbiased PU learning (UPU) \cite{du2015convex} (where unlabeled data is used as down-weighted negative examples by using a double hinge loss to remove the bias of separating positive and unlabeled data).
We perform the same ten subgraphs construction as mentioned in section \ref{experiment_engineered_PU_data} and report the mean evaluation results.
The evaluation results presented in Table \ref{eth_PU_evaluation_results}, \ref{btc_PU_evaluation_results} show that the PU learning models BA and EL perform better than the baselines for the majority of embedding methods in terms of the F1 or PUF1 for both datasets.
We also experiment with combining two types of embedding vectors.
The F1 of the combined vectors of node2vec and Poincaré representing the neighbor- and hierarchy-based representations end up with a performance of 0.946 in terms of F1 as the same performance as combining node2vec with feature engineering of nodes structure in \cite{poursafaei2021sigtran}.
From the results of the Bitcoin dataset, we can also observe an improved performance using BA and EL PU models, which illustrates the illicit node detection benefits from combining PU models in the pipeline after graph representation learning.
The combination helps in learning a dataset with potential hidden positive samples unlabeled.

\section{Conclusion}
We discussed the PU data problem in in the blockchain-based cryptocurrency transaction domain.
We also elaborate the label mechanism assumptions and the proper evaluations for blockchain transaction datasets. 
Moreover, we demonstrated the better performance of combining PU learning models into illicit node detection using graph representation learning two real-world datasets.
%
%

\bibliographystyle{ieeetran}
\bibliography{reference.bib}

\end{document}